\documentclass[conference]{IEEEtran}
\IEEEoverridecommandlockouts
\usepackage{cite}
\usepackage{amsmath,amssymb,amsfonts}
\usepackage{algorithmic}
\usepackage{graphicx}
\usepackage{textcomp}
\usepackage{xcolor}

\usepackage{pifont} % 提供 \ding 命令 
\usepackage{url} %参考文献里有一个是引用的flux的github链接，所以“引入”了 \url 命令

\def\BibTeX{{\rm B\kern-.05em{\sc i\kern-.025em b}\kern-.08em
\kern-.1667em\lower.7ex\hbox{E}\kern-.125emX}}

\DeclareRobustCommand{\authmark}[1]{\raisebox{0.85ex}[0pt][0pt]{\scriptsize\mbox{#1}}}
\DeclareRobustCommand{\authmark}[1]{\textsuperscript{#1}}

\begin{document}

\title{SimInsert: Seamless Video Object Insertion via Regional Sparse Attention Fusion \vspace{-0.5em}}

% \author{Anonymous ICME submission}
% \author{
%     \IEEEauthorblockN{\textbf{Xinyu Chen}\IEEEauthorrefmark{\text{1,2}}\IEEEauthorrefmark{1}, \textbf{Yuyi Qian}\IEEEauthorrefmark{\text{1,2}}\IEEEauthorrefmark{1},
%     \textbf{Jiang Lin}\IEEEauthorrefmark{\text{1,2}},
%     \textbf{Shenyi Wang}\IEEEauthorrefmark{\text{1,2}},
%     \textbf{Gao Wang}\IEEEauthorrefmark{\text{4}},
%     \textbf{Zhiqiu Zhang}\IEEEauthorrefmark{\text{1,2}},
%     \textbf{Jizhi Zhang}\IEEEauthorrefmark{\text{1,2}}, \\ 
%     \textbf{Mingjie Wang}\IEEEauthorrefmark{\text{5}}, 
%     \textbf{Qiang Tang}\IEEEauthorrefmark{\text{6}},
%     \textbf{Qian Wang}\IEEEauthorrefmark{\text{3}},
%     \textbf{Song Wu}\IEEEauthorrefmark{\text{3}}\IEEEauthorrefmark{2},
%     % \textbf{Zili Yi}\IEEEauthorrefmark{\text{1,2}}\IEEEauthorrefmark{2},\thanks{\IEEEauthorrefmark{1}Equal Contribution, \IEEEauthorrefmark{2}Corresponding Author. E-mail: yi@nju.edu.cn}
% }
% \IEEEauthorblockA{
%     \IEEEauthorrefmark{\text{1}}State Key Laboratory of Novel Software Technology, Nanjing University, Nanjing, China \\ 
%     \IEEEauthorrefmark{\text{2}}School of Intelligence Science and Technology, Nanjing University, Suzhou, China \\
%     \IEEEauthorrefmark{\text{3}}JIUTIAN Research, \IEEEauthorrefmark{\text{4}}Xi’an Jiaotong-Liverpool University, \\
%     \IEEEauthorrefmark{\text{5}}Zhejiang Sci-Tech University,
%     \IEEEauthorrefmark{\text{6}}The University of British Columbia 
%     }
% }

\author{
    \IEEEauthorblockN{ 
    \textbf{Xinyu Chen}\authmark{1,2,*}, 
    \textbf{Yuyi Qian}\authmark{1,2,*},
    \textbf{Jiang Lin}\authmark{1,2},
    \textbf{Shenyi Wang}\authmark{1,2},
    \textbf{Gao Wang}\authmark{4},
    \textbf{Zhiqiu Zhang}\authmark{1,2},
    \textbf{Jizhi Zhang}\authmark{1,2}, \\ 
    \textbf{Mingjie Wang}\authmark{5}, 
    \textbf{Qiang Tang}\authmark{6},
    \textbf{Qian Wang}\authmark{3},
    \textbf{Song Wu}\authmark{3,$\dagger$},
    \textbf{Zili Yi}\authmark{1,2,$\dagger$},
    \thanks{\IEEEauthorrefmark{1}Equal Contribution, \IEEEauthorrefmark{2}Corresponding Author. E-mail: yi@nju.edu.cn}
    }
    \IEEEauthorblockA{
    \authmark{1}State Key Laboratory of Novel Software Technology, Nanjing University, Nanjing, China \\ 
    \authmark{2}School of Intelligence Science and Technology, Nanjing University, Suzhou, China \\
    \authmark{3}JIUTIAN Research, \authmark{4}Xi’an Jiaotong-Liverpool University, \\
    \authmark{5}Zhejiang Sci-Tech University,
    \authmark{6}The University of British Columbia
    }
}

\maketitle

% \begingroup
% \renewcommand{\thefootnote}{}
% \footnotetext{\authmark{*}Equal Contribution, \authmark{$\dagger$}Corresponding Author. E-mail: yi@nju.edu.cn}
% \endgroup
% \setcounter{footnote}{0}

% \vspace{-12pt}

\begin{abstract}

% cxy 1224

Video object insertion requires ensuring spatio-temporal coherence and interactive realism, extending far beyond simple content placement. However, current approaches are often hindered by a reliance on explicit motion engineering or resource-intensive retraining, restricting their flexibility and generalization. To bridge this gap, we present \textit{SimInsert}, a training-free paradigm that efficiently decouples the task into intuitive single-frame editing and semantic motion description. By harnessing the robust generative priors of image-to-video diffusion models, SimInsert propagates edits temporally, strictly preserving background invariance while enabling plausible, text-driven interactions between the inserted object and the dynamic environment. Our approach hinges on non-invasive guidance mechanisms that enforce structural consistency, facilitate seamless boundary fusion, and counteract the fidelity drift that typically accumulates during the denoising trajectory. Extensive quantitative experiments validate our efficacy: SimInsert surpasses state-of-the-art methods with an 18.8\% gain in PSNR, 20.1\% in SSIM, and a 44.1\% decrease in LPIPS, offering a streamlined solution for high-fidelity video editing.

% *CRITICAL: Do Not Use Symbols, Special Characters, Footnotes, 
% or Math in Paper Title or Abstract. The abstract should contain about 100 to 150 words, and should be identical to the abstract text submitted electronically. 
\end{abstract}

% todo：keywords？写什么

\begin{figure*}[bth]
	 \centering{
		\includegraphics[width=\linewidth]{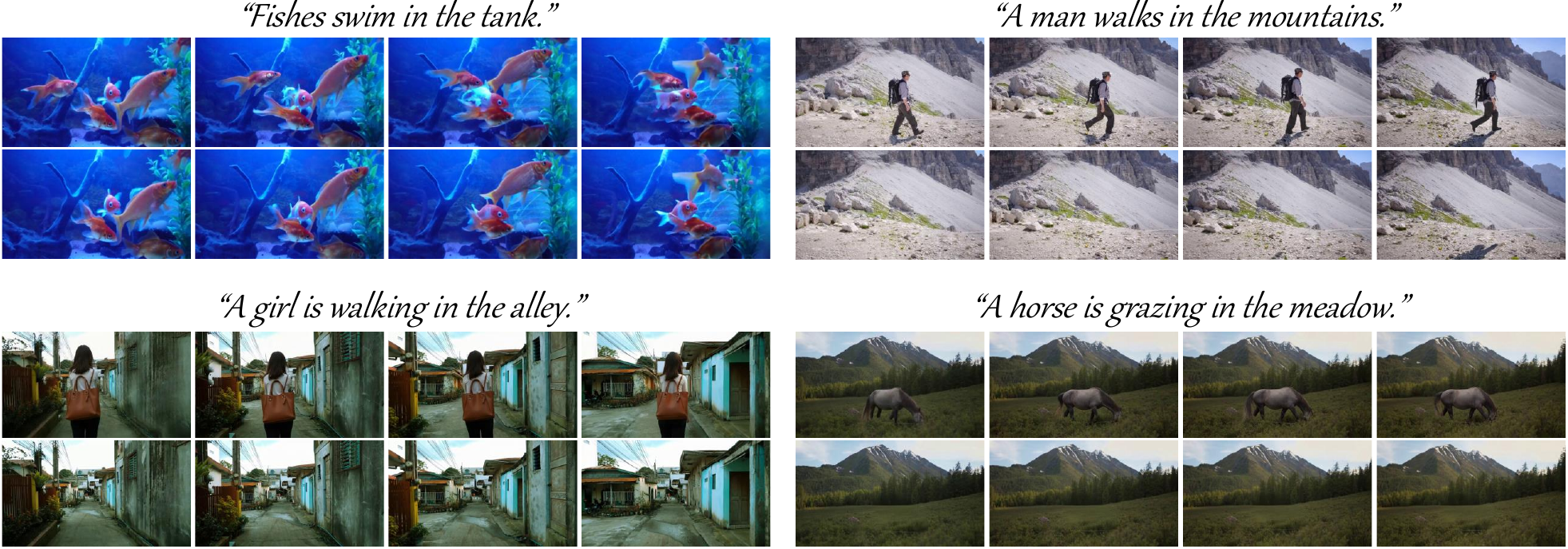}

	  \caption{\label{Qualitative Results}
Qualitative results of SimInsert. The top row displays the edited videos with the inserted objects, while the bottom row shows the corresponding original source videos. These examples illustrate our method's ability to achieve seamless integration with consistent motion while strictly preserving background fidelity.
}
}
\end{figure*}
\vspace{-10pt}

\section{Introduction}

Video editing~\cite{liu_video-p2p_2023,wei_dreamvideo_2023,gu_videoswap_2023,hu_token_2024} plays a central role in digital storytelling, entertainment, and creative expression. Unlike image editing~\cite{brooks_instructpix2pix_2023,tan_ominicontrol_2024}, which operates in a purely spatial domain, video editing necessitates reasoning over time, ensuring that modifications remain coherent across a sequence. Designing such edits is non-trivial: users must account not only for frame-wise changes but also for temporal evolution. Manual editing is tedious and impractical, particularly when inserting new content with complex dynamics.

To mitigate labor-intensive manual intervention, recent methods attempt to automate temporal propagation in different ways. \textit{Prompt-bound approaches} such as Prompt-to-Prompt~\cite{hertz_prompt--prompt_2022}, FateZero~\cite{qi_fatezero_2023}, and Re-Attentional Control~\cite{wang_re-attentional_2024} manipulate existing objects in video using textual prompts and cross-attention guidance. However, they are fundamentally limited to reinterpreting content already present in the scene—they cannot insert new objects or enable spatial freedom. 

\textit{Trajectory-driven methods} (e.g., Revideo~\cite{mou_revideo_2024}, VideoAnydoor~\cite{tu_videoanydoor_2025}) enable insertion but demand labor-intensive inputs like keypoints or bounding boxes, forcing users to manually ``design'' motion.
% Finally, motion-copying frameworks such as MVOC~\cite{wang2024mvoc} and ObjectMover~\cite{yu2025objectmover} bypass motion design entirely by inheriting trajectories from source videos. This approach is inherently inflexible and often incompatible with new environments—the copied motion may conflict with the spatial layout, lighting, or dynamics of the target scene, resulting in unnatural integration.
Finally, \textit{motion-copying frameworks} (e.g., MVOC~\cite{wang2024mvoc}) inherit trajectories from source videos, often resulting in unnatural integration due to lighting or perspective mismatches with the target scene. Consequently, a gap remains for a method that achieves high-fidelity insertion without requiring explicit motion engineering or resource-intensive training.

To bridge this gap, we present \textbf{SimInsert}, a training-free framework that fundamentally reformulates insertion by decoupling the task into two intuitive stages: \textit{spatial definition} (via flexible first-frame editing) and \textit{temporal extrapolation} (via prompt-guided motion propagation). Unlike rigid trajectory-based methods that treat objects as independent entities moving across a static canvas, SimInsert leverages the holistic generative capabilities of image-to-video (I2V) diffusion models~\cite{yang_cogvideox_2024,wan2025wan}. This allows us to synthesize the inserted object's motion and structure directly from a static edit and a prompt, ensuring fully harmonized interactions. By utilizing the model's learned physics and motion priors, SimInsert automatically resolves complex spatiotemporal dependencies—such as occlusion handling and shadow casting—without requiring manual guidance signals, 3D geometry constraints, or model fine-tuning.

However, implementing this propagation process entails two core challenges:(i) background integrity---strictly preserving the fidelity of unedited content in the original video, and (ii) interactive coherence---ensuring the inserted object interacts seamlessly with the surrounding scene dynamics.
To address the first, we propose \textbf{Regional Attention Clone}, an inversion-free mechanism that injects value matrices from a reference reconstruction path into the edited path, thereby anchoring the background content. 
To address the second, we introduce \textbf{Sparse Attention Fusion}, a strategy that stochastically blends attention scores from both the reconstruction and edited paths. This allows the generated object to perceive and react to the original scene context, ensuring smooth integration at the boundaries. 
Finally, to counteract the inevitable quality degradation in preserved regions, we incorporate a \textbf{Latent Refresh} mechanism. This module dynamically corrects cumulative noise prediction errors during denoising, ensuring that the background remains clear and consistent throughout the sequence.

Extensive experiments demonstrate that SimInsert outperforms state-of-the-art methods, offering superior temporal consistency and background preservation. Our contributions are summarized as follows:
\begin{itemize}
    \item We propose \textbf{SimInsert}, a novel training-free framework that integrates single-frame editing with I2V motion propagation. It achieves an 18.8\% improvement in PSNR, 20.1\% in SSIM, and a 44.1\% reduction in LPIPS over existing approaches.
    \item We introduce a \textbf{Regional Attention Clone} mechanism that preserves background content by injecting spatially aligned attention from a reconstruction path, eliminating the need for complex inversion.
    \item We develop a \textbf{Sparse Attention Fusion} strategy that enhances semantic coherence, enabling the inserted object to interact seamlessly with the original video context.
    \item We propose a \textbf{Latent Refresh} mechanism that mitigates background degradation by rectifying accumulated denoising drift, ensuring high-fidelity results.
\end{itemize}

\label{sec:intro}

\vspace{-5pt}
\section{Related Work}

% ==================================
% cxy 1225
\subsection{Prompt-Guided Video Editing}
Prompt-aligned editing approaches, such as Prompt-to-Prompt~\cite{hertz_prompt--prompt_2022}, FateZero~\cite{qi_fatezero_2023}, and Re-Attentional Control~\cite{wang_re-attentional_2024}, achieve controllable edits by manipulating cross-attention maps within diffusion models. While effective for style transfer or replacing existing subjects, these methods are fundamentally constrained by the layout of the source video. They rely on preserving the spatial structure of original attention maps, which restricts their ability to generate new structural content or insert objects into empty regions—a capability central to our video insertion task.

\vspace{-4pt}

\subsection{Trajectory-Based Object Insertion}
To introduce new content, several methods rely on explicit motion guidance. Revideo~\cite{mou_revideo_2024} and VideoAnydoor~\cite{tu_videoanydoor_2025} utilize user-drawn trajectories, bounding boxes, or keypoints to direct the generation of inserted objects. While these methods offer spatial control, they impose a significant manual burden: users must envision and precisely draw valid 3D trajectories on a 2D screen. This process is often unintuitive and prone to trial-and-error, as identifying a trajectory that inherently matches the scene's perspective and depth requires expert judgment.
\vspace{-5pt}

\subsection{Motion Transfer and Composition}
An alternative to manual design is borrowing motion from external videos. Frameworks like MVOC~\cite{wang2024mvoc} and ObjectMover~\cite{yu2025objectmover} insert objects by transferring trajectories from a reference clip to the target scene. Although this bypasses manual annotation, it creates a new challenge: contextual mismatch. The borrowed motion often conflicts with the target video's camera movement, spatial layout, or lighting conditions, leading to ``floating'' objects or unnatural occlusions that disrupt visual realism.
\vspace{-5pt}
\subsection{Training-Free I2V Propagation}
Recent advances in image-to-video (I2V) diffusion models have enabled training-free editing by propagating first-frame alterations temporally~\cite{ceylan_pix2video_2023,ku_anyv2v_2024,ouyang_i2vedit_2024}. These methods typically employ cross-frame attention or optical flow guidance to maintain consistency. However, applying them directly to object insertion remains problematic. They lack explicit background preservation mechanisms, often causing unedited regions to drift or hallucinate details over time. Furthermore, they offer limited control over the interaction between the new object and the scene, relying on stochastic priors rather than semantic guidance. Unlike these approaches, \textbf{SimInsert} integrates text-driven motion synthesis with robust attention-based mechanisms to ensure both background fidelity and coherent object-scene interaction.
% ==================================

\section{Method}

\begin{figure*}[bth]
	 \centering{
		\includegraphics[width=1.\linewidth ]% {framework.pdf}
        {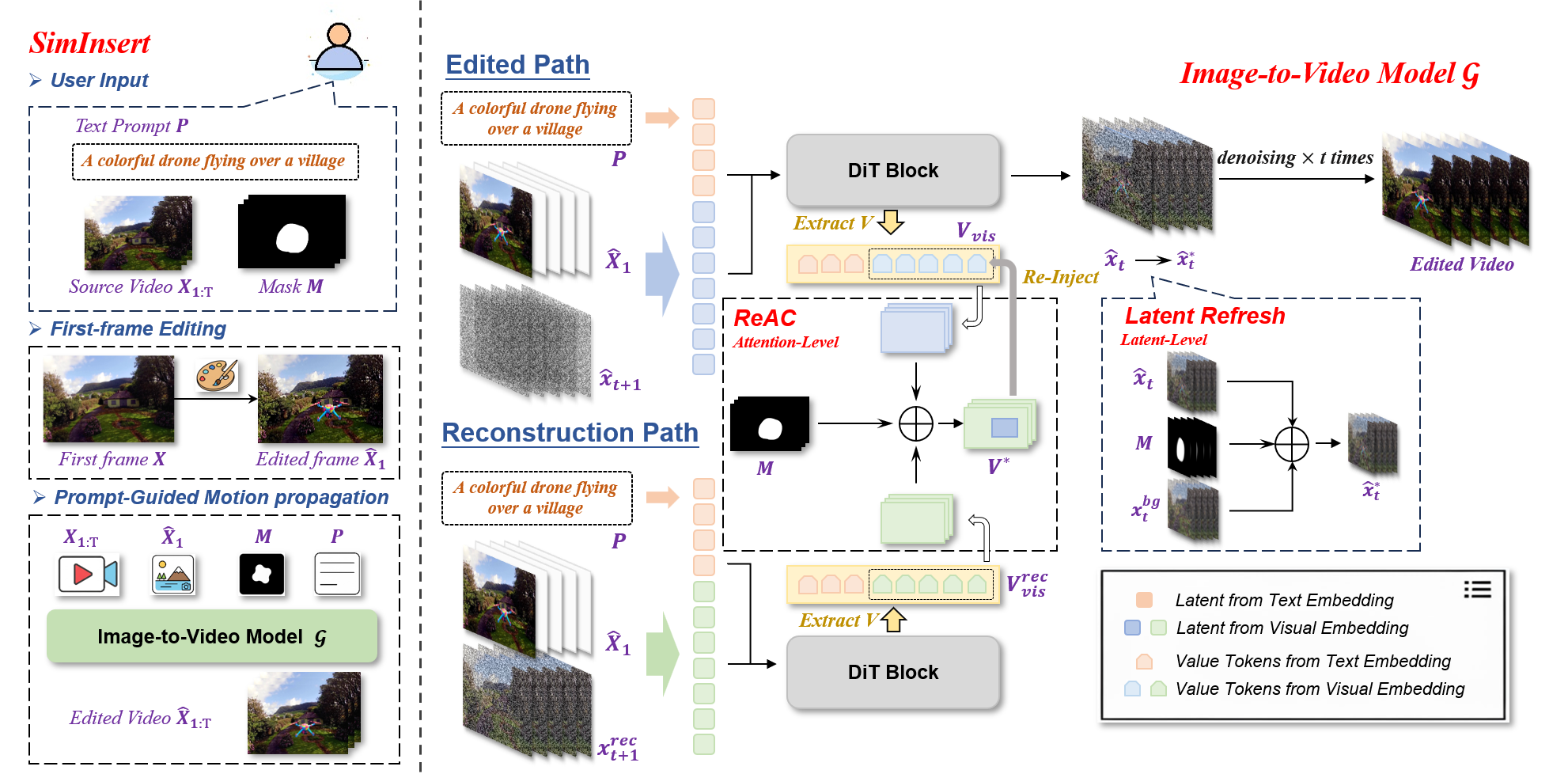}

	  \caption{\label{fig:background control}
Overview of the \textbf{SimInsert} framework. The pipeline integrates first-frame editing, prompt-guided motion propagation, and three core guidance mechanisms—Regional Attention Clone (ReAC), Sparse Attention Fusion, and Latent Refresh—into a pretrained Image-to-Video diffusion model. This architecture enables seamless video object insertion and background preservation without requiring training or manual trajectory design.}
}
\end{figure*}

As depicted in Fig.~\ref{fig:background control}, our goal is to achieve flexible video object insertion by editing only the first frame and using a natural language prompt to define the motion of the inserted content. Formally, given an input video $\mathbf{X}_{1:T}$ with $T$ frames, the user provides three inputs: (1) an edited first frame $\hat{\mathbf{X}}_1$ containing the new object, (2) a text prompt $\mathbf{P}$ describing the desired motion, and (3) a binary mask $\mathbf{M}$ indicating the region to be preserved (background) versus the region to be edited.

The objective is to synthesize a new video sequence $\hat{\mathbf{X}}_{1:T}$ that propagates the edited content from $\hat{\mathbf{X}}_1$ temporally while maintaining the fidelity of the unedited background from $\mathbf{X}_{1:T}$. We frame this as a conditional generation process using a pretrained Image-to-Video (I2V) diffusion model $\mathcal{G}$. To ensure both structural consistency and interactive realism, we adopt a “dual-path strategy”, comprising a \textit{Reconstruction path} (processing the original video) and a \textit{Edited path} (synthesizing the edited video). The process can be formulated as:
\begin{equation}
   \hat{\mathbf{X}}_{1:T} = \mathcal{G}(\hat{\mathbf{X}}_1, \mathbf{X}_{1:T}, \mathbf{P}, \mathbf{M}), 
\end{equation}
where $\mathbf{X}_{1:T}$ serves as a structural reference to guide the generation.

This formulation effectively decouples the task into \textit{content definition} (via $\hat{\mathbf{X}}_1$) and \textit{motion synthesis} (via $\mathbf{P}$). However, relying solely on the I2V prior introduces two main challenges: (i) Background Integrity, ensuring unedited regions remain pixel-aligned with the original video, and (ii) Interactive Coherence, ensuring the inserted object blends naturally with the scene context.

% To address these challenges, we introduce three lightweight guidance mechanisms that intervene in the denoising process:
To address these challenges, we introduce a hybrid guidance strategy comprising two attention-based mechanisms for structural and semantic alignment, and one latent-space mechanism for fidelity preservation:
\begin{itemize}
    \item \textbf{Regional Attention Clone (Clone):} A mechanism that directly injects self-attention features from the Reconstruction path into the background regions of the Edited path, locking in visual fidelity.
    \item \textbf{Sparse Attention Fusion (Fusion):} A strategy that blends cross-attention maps between path, allowing the inserted object to semantically "attend" to the original environment for seamless integration.
    \item \textbf{Latent Refresh (Refresh):} A correction module that periodically resets the background latents using the forward-noised original video, counteracting the accumulation of prediction errors over time.
\end{itemize}

\vspace{-5pt}
\subsection{Regional Attention Clone}
To preserve the unedited regions of the video during propagation, we introduce a regional attention clone mechanism that selectively injects attention information from the original video into the edited path. The goal is to enforce content preservation in the background while allowing new content to appear and evolve in the edited region.

Our approach operates by modifying the self-attention layers of the underlying transformer model. As illustrated in Fig.~\ref{fig:background control}, we use a dual-path setup during each sampling step: an \textit{edited path}, which processes the edited first-frame latent, and a \textit{reconstruction path}, which simulates the original video by resetting the latent at each timestep via forward linear interpolation. Let $\mathbf{x}_0$ denote the latent of the original unedited frame. The latent at timestep $t$ in the reconstruction path is determined by the flow matching dynamics:
\begin{equation}
 \mathbf{x}_t^{\text{rec}} = (1 - t)\mathbf{x}_0 + t\boldsymbol{\epsilon}
\end{equation}
where $\boldsymbol{\epsilon} \sim \mathcal{N}(\mathbf{0}, \mathbf{I})$ is the standard Gaussian noise, and $t \in [0,1]$ represents the timestep (with $t=1$ corresponding to pure noise). This expression reflects the linear probability path characteristic of flow matching models, replacing the traditional variance-preserving schedule.

In the self-attention mechanism, the input features are projected into query $Q$, key $K$, and value $V$ matrices. For text-to-video models, the value matrix $V$ typically concatenates text condition features and visual features along the sequence dimension: $V = [V_{\text{text}}, V_{\text{vis}}]$.
The standard attention operation is:
\begin{equation}
\text{Attn}(Q, K, V) = \text{softmax}\left(\frac{QK^\top}{\sqrt{d}}\right) \cdot [V_{\text{text}}, V_{\text{vis}}]
\end{equation}
To preserve the background, we extract the visual value matrix $V^{\text{rec}}_{\text{vis}}$ from the reconstruction path, which encodes the authentic spatiotemporal context of the original video. We then construct a modified visual value matrix $V^*_{\text{vis}}$ for the edited path by selectively replacing tokens in the unedited region (indicated by mask $\mathbf{M}$, where $1$ denotes the edited region and $0$ the background):
\vspace{-4pt}
\begin{equation}
 V^*_{\text{vis}} \leftarrow  \mathbf{M} \odot V_{\text{vis}} + (1 - \mathbf{M}) \odot V^{\text{rec}}_{\text{vis}}
\end{equation}
Here, $\odot$ denotes element-wise multiplication with broadcasting. The final value matrix used in the edited path becomes $V^* = [V_{\text{text}}, V^*_{\text{vis}}]$. This operation ensures that while the query $Q$ comes from the edited path (driving the generation), the background context it attends to is anchored to the original video's representation.

\begin{figure}[bth]
	 \centering{
		\includegraphics[width=\linewidth]%{sparse_fusion.pdf}% {pdf/sparse control.pdf}
        {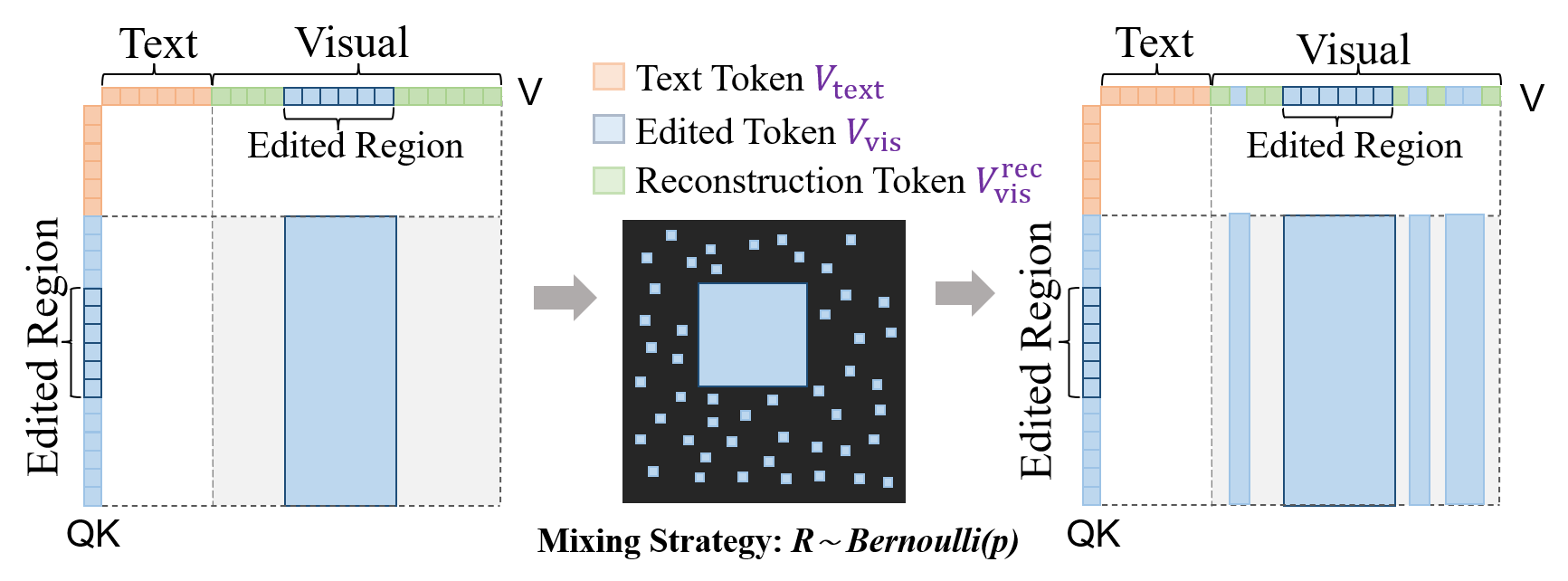}
	  \caption{\label{fig: Sparse Control Mechanism}
\textbf{Sparse Attention Fusion }mechanism. Left: attention patterns before fusion, showing limited cross-path interactions. Center: randomly sampled sparse fusion pattern. Right: post-fusion attention map, with improved blending of original and edited regions, yielding smoother spatial and temporal coherence.
}
}
\end{figure}

\vspace{-6pt}
\subsection{Sparse Attention Fusion}

While Regional Attention Clone effectively  preserves background content, simply replacing value tokens can introduce a representational mismatch. The query tokens $Q$ in the edited path are derived from a modified latent distribution, whereas the injected values $V^{\text{rec}}_{\text{vis}}$ come from the original distribution. This discrepancy can cause boundary artifacts, where the inserted object fails to interact realistically with the “frozen” background.

To mitigate this, we propose a sparse attention fusion mechanism that softly blends the generation and reconstructed value tokens within the unedited region. Let \( \mathbf{R} \in \{0,1\}^N \) be a binary mask sampled from a Bernoulli distribution with retention probability \( p \). The fused value matrix is defined as:
\vspace{-5pt}
% \begin{equation}
% V^*_{vis} = \mathbf{M} \odot V_{vis} + (1 - \mathbf{M}) \odot \left[ \mathbf{R} \odot V_{vis} + (1 - \mathbf{R}) \odot V^{\text{rec}}_{vis} \right]
% \label{eq:sparse_fusion}
% \end{equation}

\begin{equation}
V^*_{\text{vis}} \leftarrow  \mathbf{M} \odot V_{\text{vis}} + (1 - \mathbf{M}) \odot \left[ \mathbf{R} \odot V_{\text{vis}} + (1 - \mathbf{R}) \odot V^{\text{rec}}_{\text{vis}} \right]
\label{eq:sparse_fusion}
\end{equation}

Sparse fusion is applied only within the unedited region. For each token \( j \) where \( \mathbf{M}_j = 0 \), we sample a binary mask \( \mathbf{R}_j \sim \text{Bernoulli}(p) \) to retain the generation value token \( V_j \), or replace it with its reconstruction counterpart \( V_j^{\text{rec}} \). Edited regions (\( \mathbf{M}_j = 1 \)) remain untouched.

This mechanism yields two complementary effects. First, it restores representational alignment over time. By sparsely retaining edited-path tokens as anchors, it maintains compatibility between edited queries and attended content. This stochastic exposure progressively aligns the edited representation with the background latent space—termed \textit{consistent guidance accumulation}—resolving latent mismatch without disrupting structural evolution.
Second, sparse fusion preserves semantic continuity across space. Unlike hard replacement that creates boundary artifacts, the presence of edited-path tokens ensures a shared latent lineage within local neighborhoods. This smooths transitions across mask boundaries, mitigating visual seams and enabling coherent integration.
Rather than overwhelming the model with opposing signals, sparse fusion softly mediates between them to facilitate natural object-scene interaction. This directly alleviates a critical challenge in localized video editing: seamlessly integrating inserted content with the existing background while ensuring coherence throughout the generative process.

\subsection{Latent Refresh for High‑Fidelity Backgrounds}

%  ====================
%  ========cxy 1228====

Although Regional Attention Clone preserves structural layout, we observe that the visual fidelity of the background often degrades over long generation sequences. This degradation is intrinsic to the generative process, distinct from the interaction issues addressed by Sparse Fusion.

The core cause lies in cumulative numerical drift. In the attention mechanism, the query $Q$ originates from the edited path, while the value $V^{\text{rec}}$ is sourced from the reconstruction path. Since the edited latents diverge slightly from the original trajectory, this cross-path discrepancy introduces minor deviations in the attention output. Because the flow matching integration (i.e., the ODE solver) is an iterative process, these small errors accumulate at each step, eventually manifesting as texture blurring, color shift, or residual noise in the unedited regions.

To resolve this, we introduce a \textbf{Latent Refresh} mechanism that periodically resets the background latents to their ideal trajectory. Since our backbone employs Flow Matching~\cite{lipman2023flowmatchinggenerativemodeling}, the forward process is defined as a linear interpolation between the data and noise distributions. At each timestep $t$, we explicitly compute the noised-background latent $\mathbf{x}_t^{\text{bg}}$ from the original video $\mathbf{x}_0$ and the initial noise $\boldsymbol{\epsilon}$:
\vspace{-4pt}
\begin{equation}
 \mathbf{x}_t^{\text{bg}} = (1 - t) \mathbf{x}_0 + t \boldsymbol{\epsilon}, \quad \boldsymbol{\epsilon} \sim \mathcal{N}(\mathbf{0}, \mathbf{I})
\end{equation}
Here, we assume $t \in [0, 1]$ represents the noise level ($t=0$ for clean data). Crucially, we use the same noise instance $\boldsymbol{\epsilon}$ sampled for the current generation process to ensure strictly aligned trajectories. We then blend this noised-background  latent into the current edited latent $\hat{\mathbf{x}}_t$ before the vector field prediction:
\vspace{-4pt}
\begin{equation}
 \hat{\mathbf{x}}_t^* \leftarrow \mathbf{M} \odot  \hat{\mathbf{x}}_t + (1 - \mathbf{M}) \odot \mathbf{x}_t^{\text{bg}}
\end{equation}
\vspace{-4pt}
This operation effectively ``refreshes'' the background region, enforcing pixel-level fidelity to the source video. By anchoring the background state to the exact linear probability path of the original video, we neutralize the accumulated numerical errors, ensuring that the background remains crisp and consistent throughout the sequence.

\section{Experiments}

\begin{table*}[htbp]
\caption{Quantitative comparison on the benchmark. SimInsert achieves superior performance across all metrics.
Note that VFID is omitted for FateZero (8 frames) and ConsistI2V (16 frames) due to frame count discrepancies.}
\label{tab:quantitative}
\centering
\small
\begin{tabular}{lcccccc}
\hline
Metric & $\uparrow$ PSNR & $\uparrow$ SSIM & $\downarrow$ LPIPS &
$\downarrow$ VFID & $\uparrow$ CLIP-I & $\uparrow$ CLIP-T \\
\hline
pix2video  & 28.96 & 0.7219 & 0.3852 & 1479.67 & 0.9803 & 0.2789 \\
FateZero   & 29.18 & 0.6345 & 0.3835 & --      & 0.9737 & 0.2499 \\
ConsistI2V & 30.52 & 0.5903 & 0.4721 & --      & 0.9788 & 0.2837 \\
AnyV2V     & 29.75 & 0.6815 & 0.2630 & 1240.54 & 0.9814 & \textbf{0.2867} \\
Ours       & \textbf{36.26} & \textbf{0.8671} & \textbf{0.1471} &
             \textbf{1062.92} & \textbf{0.9923} & 0.2825 \\
\hline
\end{tabular}
\end{table*}

\begin{figure*}[htb]
    \centering
    \includegraphics[width=0.85\linewidth]{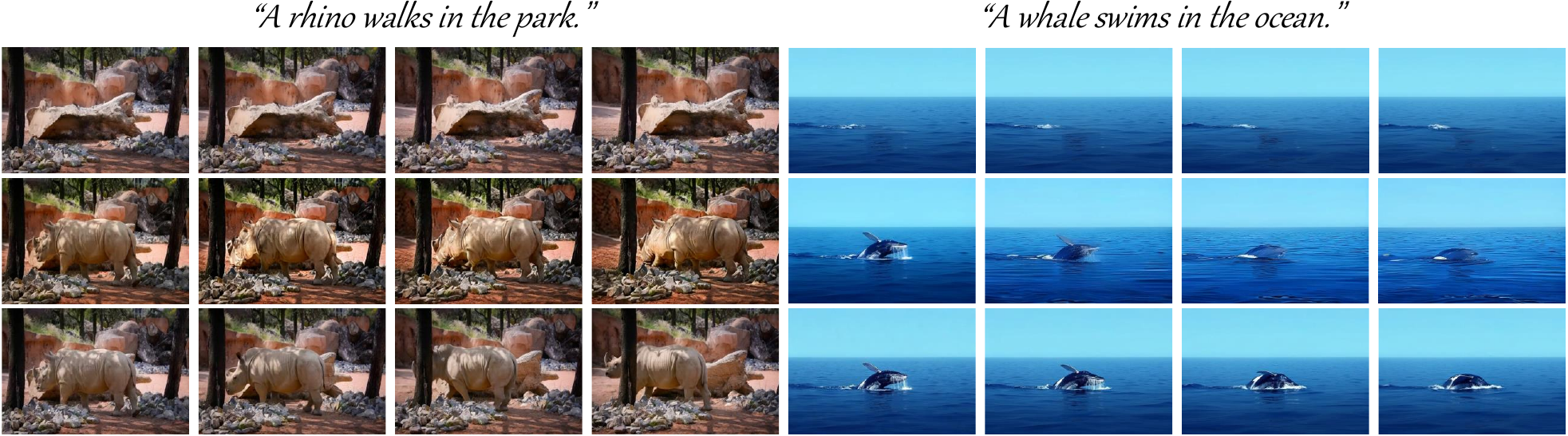}
    \caption{Qualitative comparison. Top: Original/Input video. Middle: The strongest baseline, AnyV2V. Bottom: SimInsert (Ours).}
    \label{fig:qualitative}
\end{figure*}
\vspace{-6pt}

We evaluate SimInsert across synthetic and real-world scenarios to assess its ability to produce coherent, high-fidelity object insertions under diverse and challenging conditions. \textit{Due to space constraints, detailed ablation studies, extended applications, and additional visual results are provided in the Supplementary Material.}

\subsection{Experimental Setup}

\textbf{Dataset Construction.} Since no existing dataset is designed for \textit{free-form, text-guided video object insertion}, we constructed a comprehensive benchmark to enable consistent evaluation. The dataset comprises a diverse mix of real-world and synthetic sequences. The real-world subset is drawn from the DAVIS dataset~\cite{ponttuset20182017davischallengevideo}, selected for its high-quality footage and challenging motion dynamics. 
% For the synthetic subset, we remove objects from the first frame using ZITS~\cite{dong_incremental_2022}, generate a clean background video, and then perform insertion using the original object mask. 
To adapt these sequences for insertion task, we employ MiniMax-Remover~\cite{zi2025minimax} to erase the primary subjects from the video clips, serving as the background reference. The original, unaltered first frames (containing the subjects) are then utilized as the edited inputs to initiate the propagation.
For the synthetic subset, we first generate clean base video sequences and then employ FLUX Kontext~\cite{flux2024} to insert objects into the first frame, thereby creating the edited inputs.
All videos are standardized to 49 frames at 480p resolution for evaluation consistency, though SimInsert supports arbitrary lengths permitted by the base model.

\textbf{Backbones and Tools.} In this evaluation, SimInsert is primarily built on Wan2.1~\cite{wan2025wan}, but its design is model-agnostic and can be applied to any Image-to-Video diffusion backbone, such as CogVideoX5B-I2V~\cite{yang_cogvideox_2024} or LTXVideo~\cite{hacohen2024ltx} (Please refer to Supplementary Material Section IV for the experimental results of these models). First-frame edits are created using off-the-shelf image editing tools, such as Flux Kontext~\cite{flux2024}.

\textbf{Hardware.} All experiments are run on NVIDIA A6000 GPUs (48\,GB). For Wan~2.1 variant, it requires either two A6000 GPUs or a single GPU with greater VRAM. Runtime is roughly 1.5--2$\times$ the  I2V inference time, depending on the backbone.

\textbf{Baselines.} We benchmark against Pix2Video~\cite{ceylan_pix2video_2023}, FateZero~\cite{qi_fatezero_2023}, ConsistI2V~\cite{ren_consisti2v_2024}, and AnyV2V~\cite{ku_anyv2v_2024}. FateZero is evaluated using its default 8-frame setting, as increasing the frame count would dramatically raise VRAM requirements. Because this shorter sequence length distorts the VFID measurement, we exclude FateZero from VFID comparisons.

\subsection{Quantitative Results}

We evaluate \textbf{SimInsert} against recent video editing and insertion methods using six metrics: PSNR, SSIM, LPIPS, VFID, CLIP-I, and CLIP-T. PSNR, SSIM, and LPIPS are calculated only on the unedited regions to assess background fidelity preservation. VFID (Video Fréchet Inception Distance) reflects overall perceptual realism and how closely generated videos match the distribution of real videos. CLIP-I measures frame-to-frame semantic consistency, serving as an indicator of temporal continuity, while CLIP-T evaluates how well the generated content aligns with the guiding prompt.

All metrics are reported over the full dataset. 
As shown in Table~\ref{tab:quantitative}, SimInsert consistently achieves the highest scores across nearly all metrics, demonstrating sharper backgrounds, stronger temporal continuity, and closer alignment with the guiding edits, with an \textbf{18.8\% improvement in PSNR, 20.1\% in SSIM, and a 44.1\% reduction in LPIPS} over the best-performing baselines.

\subsection{Qualitative Results}

We qualitatively compare SimInsert with prior works to demonstrate its advantages in background preservation and seamless integration. Fig.~\ref{fig:qualitative} highlights a side-by-side comparison with AnyV2V, the strongest baseline. We focus on this comparison as other methods frequently exhibit severe artifacts or temporal instability that preclude meaningful visual analysis. As observed, while AnyV2V generally maintains background structure, the inserted objects often suffer from \textit{temporal degradation}: moving erratically, remaining static, or collapsing into incoherent textures as generation progresses. This reveals a failure to maintain object identity over time.
In contrast, SimInsert produces stable backgrounds and naturally moving objects. The visual evidence confirms that our Regional Attention Clone effectively prevents background distortions, while Sparse Attention Fusion eliminates boundary seams. Furthermore, the Latent Refresh mechanism ensures that the background remains crisp throughout the sequence, avoiding the "washed-out" effect common in long-horizon generation.

\vspace{-5pt}

\section{Conclusion}
We present \textbf{SimInsert}, a training-free framework for video object insertion that can transform any I2V diffusion model to perform robust local editing by combining first-frame editing with prompt-guided motion propagation. By introducing regional attention clone for background preservation and sparse attention fusion for gradual alignment between edited and unedited regions, SimInsert addresses the chronic incoherence that typically hinders localized video editing, transforming fragmented generations into seamless integrations. Our results demonstrate that lightweight interventions in attention representations can achieve temporal and spatial coherence previously thought to require retraining or heavy supervision—pointing toward a new paradigm of controllable, insight-driven generative video editing.

\section*{Acknowledgment}
This work was supported by the National Natural Science Foundation of China (Grant No. 62406134), Jiangsu Provincial Science \& Technology Major Project (Grant No. BG2024042),  the Suzhou Key Technologies Project (Grant No. SYG2024136) and the Nanjing University-China Mobile Communications Group Co. Ltd. Joint Institute.

\bibliographystyle{IEEEtran}
\bibliography{main}

% \vspace{12pt}
% \color{red}

\end{document}